\pdfoutput=1

\documentclass[11pt]{article}
\usepackage[final]{ACL2023}

\usepackage{times}
\usepackage{latexsym}
\usepackage{amsmath} 
\usepackage{longtable}
\usepackage{supertabular,booktabs}
\usepackage{graphicx}
\usepackage{multicol}
\usepackage{multirow}
\usepackage[T1]{fontenc}

\usepackage[utf8]{inputenc}

\usepackage{microtype}

\usepackage{inconsolata}
\usepackage{tabularx} 
\usepackage{array}    

\usepackage{algorithm} 
\usepackage{algpseudocode} 
\newcommand{\ours}{P-FOLIO }

\title{P-FOLIO: Evaluating and Improving Logical Reasoning with Abundant Human-Written Reasoning Chains}

\author{
 Simeng Han$^{1}$ 
   \quad \textbf{Aaron Yu}$^{2}$ 
  \quad \textbf{Rui Shen}$^{2}$ 
 \quad \textbf{Zhenting Qi}$^{2}$ \\
\quad \textbf{Martin Riddell}$^{1}$ 
 \quad \textbf{Wenfei Zhou}$^{3}$ 
 \quad \textbf{Yujie Qiao}$^{1}$ 
  \quad \textbf{Yilun Zhao}$^{1}$ \\
  \quad \textbf{Semih Yavuz}$^{4}$  
  \quad \textbf{Ye Liu}$^{4}$  
  \quad \textbf{Shafiq Joty}$^{4}$ 
  \quad \textbf{Yingbo Zhou}$^{4}$ \\
 \quad \textbf{Caiming Xiong}$^{4}$ 
  \quad \textbf{Dragomir Radev}$^{1,4}$
 \quad \textbf{Rex Ying}$^{1}$ 
 \quad \textbf{Arman Cohan}$^{1}$  \\ 
  $^1$Yale University, 
  $^2$Harvard University,
  $^3$NVIDIA,
  $^4$Salesforce Research 
 }

\begin{document}
\maketitle

\begin{abstract}
Existing methods on understanding  the capabilities of LLMs in logical reasoning rely on binary entailment classification or synthetically derived rationales, which are not sufficient for proper investigation of model's capabilities. We present \textit{P-FOLIO}, a human-annotated dataset consisting of diverse and complex reasoning chains for a set of realistic logical reasoning stories also written by humans. P-FOLIO is collected with an annotation protocol that facilitates humans to annotate well-structured natural language proofs for first-order logic reasoning problems in a step-by-step manner. The number of reasoning steps in P-FOLIO span from 0 to 20. We further use P-FOLIO to evaluate and improve large-language-model (LLM) reasoning capabilities. We evaluate LLM reasoning capabilities at a fine granularity via single-step inference rule classification, with more diverse inference rules of more diverse and higher levels of complexities than previous works. Given that a single model-generated reasoning chain could take a completely different path than the human-annotated one, we sample multiple reasoning chains from a model and use pass@k metrics for evaluating the quality of model-generated reasoning chains. 
We show that human-written reasoning chains significantly boost the logical reasoning capabilities of LLMs via many-shot prompting and fine-tuning. Furthermore, fine-tuning Llama3-7B on P-FOLIO improves the model performance by 10\% or more on three other out-of-domain logical reasoning datasets. We also conduct detailed analysis to show where most poweful LLMs fall short in reasoning. We will release the dataset and code publicly.


\end{abstract}

\section{Introduction}

\begin{table*}[t!]
\centering

\label{tab:comparison}
\resizebox{\textwidth}{!}{
\begin{tabular}{@{}lcccc@{}}
\toprule
\multirow{2}{*}{\textbf{Dataset}} & \multicolumn{2}{c}{\textbf{Natural Language}} & \multicolumn{2}{c}{\textbf{Logic}} \\
 & \textbf{Text Source} & \textbf{Real-World Resource} & \textbf{\# Reasoning Step} & \textbf{\#Inference rule} \\
 \cmidrule(lr){2-3} \cmidrule(lr){4-5} 
\textbf{ProofWriter \citep{tafjord-etal-2021-proofwriter}} & Synthetic  & No  & 0 - 5 &  12 \\


\textbf{ProntoQA-OOD \citep{saparov2023testing}} &  Synthetic &  No & \texttimes &  6 \\

\textbf{\ours}    & Expert-written  & Yes & 0 - 20 & 31 \\
\bottomrule
\end{tabular}
}
\caption{Comparison of \ours and other datasets of proofs}
\label{tab:proof-data-comparison}
\end{table*}

\begin{table*}[t!]
\centering
\resizebox{\textwidth}{!}
{%
\begin{tabular}{ll}
\toprule
A FOLIO example& \\
\midrule
\textbf{NL premises} & \textbf{NL Conclusion -> Label} \\
1. There are no mansion houses in an urban area. & If the LaLaurie House is either a   \\ 
2. All skyscrapers are in urban areas. & skyscraper or a creepy haunted house, \\
3. Every creepy haunted house is a mansion house. & then it is not a mansion house. -> F\\ 
4. Every terrifying building on Halloween is a creepy haunted house. & \\ 
5. The LaLaurie House is a creepy haunted house or a terrifying Halloween building. & \\ 


\bottomrule
\end{tabular}
}
\captionof{table}{An example story in FOLIO consisting of five premises and one conclusion with the label for the conclusion. 
}
\label{tab:example}
\end{table*}

A logical reasoning story consists of a series of premises and one or more conclusions \citep{han2022folio}. The goal of logical reasoning is to determine the truth values of the conclusions based on the premises \citep{russell-norvig-2010}. However, it cannot be determined whether a machine is indeed equipped with capabilities of logical reasoning even if it is able to output the correct truth value. Generating proofs or reasoning chains that support its decision is essential for explainability. Resolution-based symbolic proofs can be generated automatically with a first-order logic solver given explicit logic structures, but they are hard to read and cannot be written in natural language in a straightforward way in most cases \citep{russell-norvig-2010}. Although studies on natural language proof generation for logical reasoning have been conducted before, they focus on synthetically generated logical reasoning datasets \cite{saparov2023testing, saha-etal-2020-prover, tafjord-etal-2021-proofwriter, saha-etal-2021-multiprover}. These logical reasoning chains contains much less natural language variation and challenging reasoning patterns than realistic logic stories \citep{han2022folio}.

 We present \textit{\ours}, a new dataset consisting of human-written proofs for an existing popular dataset of logical reasoning, FOLIO \citep{han2022folio} which consists of realistic logical reasoning stories written by humans. \ours is collected with an annotation protocol for annotators to write proofs in a step-by-step manner without using proof by contradiction or proof by case. Table~\ref{tab:example} shows an example story in FOLIO and Table~\ref{tab:reasoningchain} shows its human-written reasoning chains in P-FOLIO. 


\ours proofs are expert-written. They are more logically diverse and challenging than previous logical reasoning datasets equipped with reasoning chains. The number of proof steps in \ours spans from 0 to 20 and the proofs use diverse inference rules, containing 12 types of widely used and straightforward inference rules and another 20 types of complex inference rules. $26\%$ of a total of \textbf{1430} proofs in \ours consist of more than five reasoning steps and $4\%$ of proofs consist of more than 10 reasoning steps. We show the comparsion of P-FOLIO and the previous proof collections for logical reasoning,  ProofWriter \citep{tafjord-etal-2021-proofwriter} and ProntoQA-OOD\citep{saparov2023testing}, in Table~\ref{tab:proof-data-comparison}. Both ProofWriter and ProntoQA-OOD are synthetically constructed. ProofWriter contains at most 5 reasoning steps and ProntoQA-OOD only uses 6 types of inference rules. 

We use \ours to provide a more in-depth evaluation and to improve large-language-model reasoning capabilities. We first evaluate LLM reasoning capabilities at a fine-grained granularity via single-step inference rule classification and single-step derivation reasoning. Given that the model-generated reasoning chain could take a completely different path than the human-annotated one, we use the pass@k  \cite{Chen2021EvaluatingLL,kulal2019spoc}, a metric commonly used for evaluating code generation, for evaluating the overlap between them. 
We show that human-written reasoning chains significantly boost the logical reasoning capabilities of LLMs via many-shot prompting and fine-tuning. Furthermore, fine-tuning Llama3-7B on P-FOLIO improves the model performance by more than 10\% on three other out-of-domain logical reasoning datasets. Finally, we conduct detailed analysis to show where most poweful LLMs fall short in reasoning. 

Our contributions are as follow:
(1) We collected expert-written reasoning chains for an existing popular logical reasoning dataset. The collected reasoning chains are more logically diverse and challenging than previous logical reasoning datasets equipped with reasoning chains.
(2) We evaluate LLM-generated proofs with both automatic metrics and human evaluation. 
(3) We show that fine-tuning on our proof collection yields substantial improvements for LLM logical reasoning capabilities on both in-domain and out-of-domain test sets.
(4) We identify weaknesses of LLM reasoning based on our findings.

\section{Related Work}

\paragraph{Natural language reasoning}
Existing studies on natural language reasoning \citep{wei2022chain, creswell2022selection, bang2023multitask, prystawski2023think, yao2023tree} focus on using different prompting strategies on large language models (LLMs) for better results. Despite significant progress on the LLM prediction accuracy of the final answer, no work has been done to evaluate the correctness of the intermediate steps. In our work, we propose to evaluate intermediate steps at different levels of granularity in order to better probe into the reasoning capabilities of LLMs. 

\paragraph{Improving LLM Logical reasoning capabilities}

\citet{luo2024logiglue, ranaldi-freitas-2024-aligning} fine-tuned language models with logical reasoning data to improve logical reasoning capabilities of LLMs. Large language models have also been directly used as soft logic reasoners and a variety of prompting techniques are proposed in order to improve their performance under this paradigm \citep{wei2022chain, yao2023tree, zhou2024selfdiscover}. Using large language models as semantic parsers has also shown improvement on the reasoning performance \citep{linc, pan-etal-2023-logic} where natural language reasoning problems are first parsed into logical forms before being fed into an inference engine to output the final answer. 
 

\paragraph{Natural language proof generation}
ProofWriter \citep{tafjord-etal-2021-proofwriter} and FLD \citep{pmlr-v202-morishita23a} are logical reasoning datasets equipped with natural language proofs, however both of them are synthetically generated dataset which neither contains abundant natural language variation nor encompasses challenging logical reasoning patters. Previous studies on proof generation focus on ProofWriter\citep{pmlr-v202-morishita23a, saha-etal-2020-prover, saha-etal-2021-multiprover, yang-etal-2022-generating} and ProntoQA \citep{saparov2023testing}. LogicBench is a synthetically generated natural language QA dataset and is used for evaluating the logical reasoning ability of LLMs \citep{parmar-etal-2024-logicbench}. While FOLIO covers first-order logic and one or more inference rules are used in each example, LogicBench focuses on reasoning patterns covering propositional logic, first-order logic, and non-monotonic logic and focuses on the use of a single inference rule for each example. 

We collect proofs for FOLIO \citep{han2022folio} instead, a realistic expert-written logical reasoning dataset. Such proofs need to be written from scratch and are hard and time-consuming to write because humans need to manage both the language and reasoning complexity in the proof-writing process and manually construct many steps of reasoning. The resulting proofs contain more diverse types of inference rules and reasoning patterns in addition to containing more natural language variation and ensured semantic richness.




\section{\ours}
In this section we describe the construction and properties of our dataset \ours. 

\begin{table*}[t!]
\centering
\resizebox{\textwidth}{!}
{%
\begin{tabular}{llll}
\toprule
\textbf{P No Used} & \textbf{Derivation} & \textbf{D No. } & \textbf{Inference Rule} \\
\midrule 
1 & If there is a mansion house in the area, then it is not an urban area. & D1 & Modus Tollens \\
2 & Something that is not in an urban area is not a skyscraper. & D2 & Modus Tollens \\
3, D1 & A creepy haunted house is not in an urban area. & D3 & Hypothetical Syllogism \\
D3, 2 & A creepy haunted house is not a skyscraper. & D4 & Hypothetical Syllogism \\
4, D4 & Every terrifying Halloween building is not a skyscraper. & D5 & Hypothetical Syllogism \\
D5, 5 & The LaLaurie House is not a skyscraper or is a terrifying Halloween building. & D6 & Universal Instantiation \\
D6, D4 & The LaLaurie House is not a skyscraper or is not a skyscraper. & D7 & Universal Instantiation \\
D7 & The LaLaurie House is not a skyscraper. & D8 & Idempotence \\
5, 4 & The LaLaurie House is a creepy haunted house or a creepy haunted house. & D9 & Universal Instantiation \\
D10 & The LaLaurie House is a creepy haunted house. & D11 & Idempotence \\
D11, 3 & The LaLaurie House is a mansion house. & D12 & Universal Instantiation \\
D8, D11 & The LaLaurie House is a creepy haunted house and not a skyscraper. & D13 & Conjunction Introduction \\
D13, D12 & The LaLaurie House is a creepy haunted house and a mansion house and not a skyscraper. & D14 & Conjunction Introduction \\
\bottomrule
\end{tabular}
}
\captionof{table}{Proof written for the example story in FOLIO. ``P No'' is premise number and ``D'' stands for derivation. 
}
\label{tab:reasoningchain}
\end{table*}

\begin{table*}[ht]
\centering
\setlength{\tabcolsep}{3pt}
\small
\begin{tabular}{lp{11cm}}
\toprule
\textbf{Type} & \textbf{Rules} \\
\midrule
\textbf{Widely-used rules} & universal instantiation (UI), hypothetical syllogism (HS), Modus Ponens (MP), Modus Tollens (MT) \\
 & disjunctive syllogism (DS), disjunction introduction (DI), conjunction introduction (CI) \\
 & conjunction elimination (CE), transposition (TP), material implication (MI), existential introduction (EI) \\ 
\midrule
\textbf{Boolean Identities (Idempotent)} & "A or A" is equivalent to A; A and A is equivalent to A \\ 
\midrule
\textbf{Complex rules} & 
From $a \oplus b$, we know $a \rightarrow \lnot b$; from $a \rightarrow b$ and $b \rightarrow a$, we know $\lnot(a \oplus b)$. \\

& $\lnot b \oplus b$ is always true; $a \oplus b$ is equivalent to $ \lnot a \oplus \lnot b$
 \\ 

& From $\forall \mathrm{x}(\mathrm{A}(\mathrm{x}) \rightarrow \mathrm{C}(\mathrm{x}))$ and $\mathrm{A}(paul) \wedge B(paul)$, we know $C(paul) \wedge B(paul)$.  \\

\bottomrule
\end{tabular}
\caption{Categories of inference rules used in our annotation protocol. }
\label{tab:inference-rules}
\vspace{-1.5em}
\end{table*}

\subsection{Inference Rules}
We first define a set of inference rules that can be used for derivations of each proof step. 
\paragraph{Widely-used inference rules.} The most widely used logical reasoning inference rules include universal instantiation, hypothetical syllogism, modus ponens, modus tollens, disjunctive syllogism, conjunction introduction, conjunction elimination, transposition, disjunction introduction, material implication and existential introduction.

\paragraph{Boolean identities.} During the pilot annotation process, we found that boolean identities are needed for certain derivations. For example, if "A or A" is true then "A" is true. We therefore allow the usages of boolean identities 

\paragraph{Complex inference rules.} Some complex inference rules are intuitively correct and can also be proved logically correct with an inference engine. For example, from "A XOR B", we know that "A implies not B". We include in Table~\ref{tab:inference-rules} the different categories of inference rules used in our protocol. 

\subsection{Dataset Annotation}
We offer an annotation protocol that facilitates annotators to annotate well-structured proofs for any logical reasoning problems in a step-by-step manner. Our annotators are selected based on the the following criteria: 
1). They are college or graduate students who are either native English speakers or have near-native English proficiency. 2) They have a formal background in first-order logic, gained through either relevant coursework or self-directed study in first-order logic or semantic parsing. 3). we conducted in-person interviews with all the annotators to understand if they are motivated about completing the task before we added them into the annotator pool. All the annotators selected have expressed a keen interest in solving logical puzzles and are guaranteed to have a significant amount of time commitment. We also give the annotators several detailed tutorial sessions on how to write a proof and give detailed annotation guidelines to minimize possible ambiguities that can occur in the annotation process and make sure the annotation protocol is well-understood. A total of six annotators were selected through this process. 

\paragraph{Annotation protocol.} An annotator is presented with a logical reasoning problem consisting of a series of premises and one conclusion. For each step of a proof, we ask the annotator to write the indices of the premises used, a natural language derivation deduced from the premises used and the inference rule used for the derivation. The natural language derivation can then be used as a premise for future proof steps. We also specify that only 1 or 2 premises should be used in each step. Table \ref{tab:example} shows an example of the FOLIO dataset and Table~\ref{tab:reasoningchain} shows the entire proof annotated for the example. The proof consists of 14 steps in total with 14 natural language derivations and the final derivation is the conclusion. This annotation protocol facilitates humans to annotate well-structured proofs for first-order logic reasoning problems in a step-by-step manner and the resulting proofs are easier to read and evaluate than proofs written in pure natural language without being divided into individual steps.

We provide the detailed guidelines on proof annotation in the appendix. The guideline contains detailed instructions on how to write the proofs. Multiple examples were provided to the annotators and we conducted in-person tutorials for the annotators. We promptly ask the annotators for feedback and resolve their questions throughout the entire annotation process. We found that annotators are well-versed in the well-established inference rules and boolean identities. However, sometimes universal instantiation inferences can be wrongly identified as hypothetical syllogism. At the initial stage of annotation, annotators sometimes identified intuitively correct but logically incorrect complex rules. Therefore, we ask them to verify each complex rule with a logic solver before using it.

We asked the annotators to keep a record of the time taken for writing each proof. Most of the proofs requiring 1-5 steps take less than 20 minutes to annotate with five minutes being the average while more complicated proofs requiring six steps or more take 20 minutes to 90 minutes to annotate with 1 hour being the average since to write these proofs requires a significant amount of deliberate thinking and annotators need to go through an intermediate deduction process before writing down the proof steps. 

For first-order logical reasoning questions, multiple reasoning paths can be undertaken to derive the same conclusion. Due to the time-consuming nature of the task, only one annotator annotates each sample with one reasoning path in our annotation procedure although in our pilot annotation study we identified different annotators sometimes provide different proofs for the same example. 

\paragraph{Quality control via cross checking.} We conducted a comprehensive cross-checking process for all the proofs written by dividing the annotators into 3 groups, each comprised of 2 people. We asked the two annotators in each group to review and validate each other's work and address any identified errors. We also provided annotators with cross checking guidelines to make the process more efficient. This approach greatly reduces errors for P-FOLIO. 

\newcommand*\xor{\oplus}

\subsection{Dataset statistics}
\begin{figure}[t]
    \centering
    \includegraphics[width=7cm]{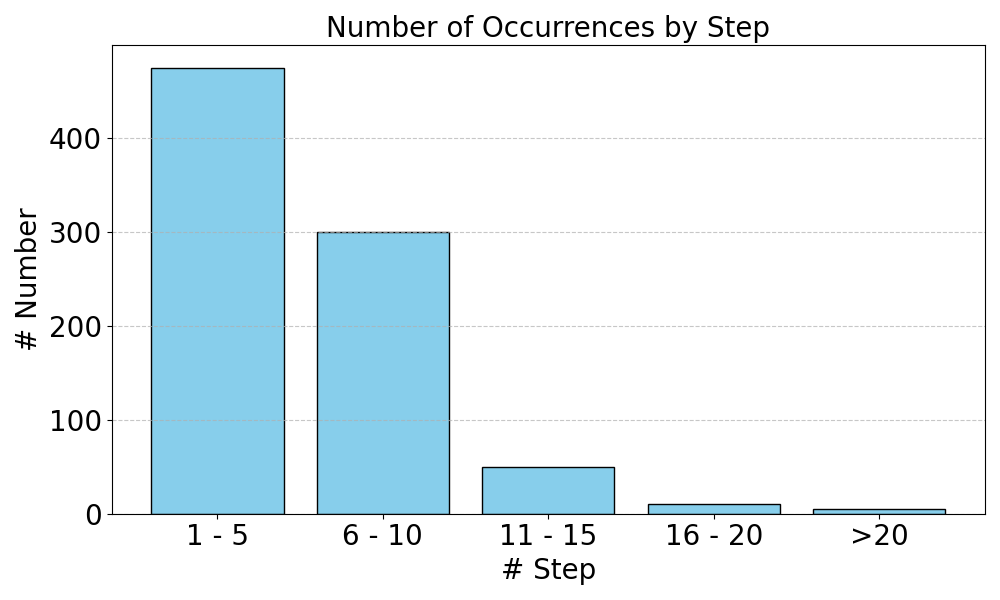}
    \caption{Distribution of the number of proof steps}
    \label{fig:proof_step_dist}
\end{figure}

\paragraph{Number of proof steps.} We show the distribution of the number of proof steps in the entire \ours dataset in Figure \ref{fig:proof_step_dist}. we have carefully validated that each of the splits follows a similar distribution with less than a 2\% difference in each split. The majority of written proofs contain five steps or less while written proofs comprised of 6-10 steps also take up a significant portion. There are 55 proofs which are comprised of 11 steps or more. This represents the portion of the dataset with the highest complexity.

\begin{figure}[t]
    \centering
    \includegraphics[width=8cm]{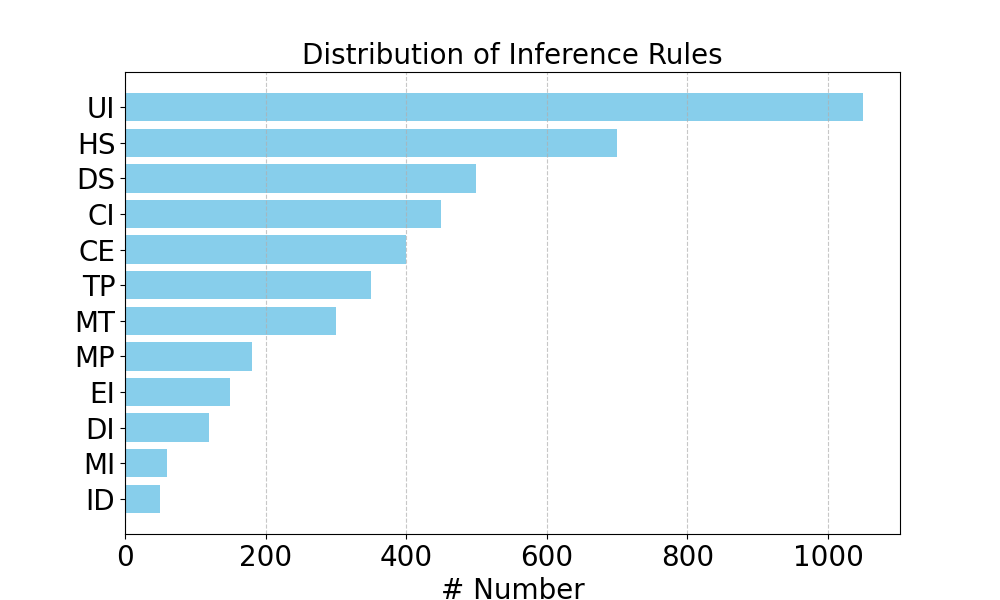}
    \caption{Distribution of widely used inference rules}
    \label{fig:inference_rule_dist}
\end{figure}

\paragraph{Inference rule distribution.} The widely-used inference rule and boolean identities distribution is shown in Figure~\ref{fig:inference_rule_dist}. Among the widely-used inference rules, universal instantiation and hypothetical syllogism are the most common inference rules in \ours while disjunctive syllogism, conjunction introduction, conjunction elimination and transposition also appeared a large number of times. Although Modus ponens, Modus tollens, existential introduction, disjunction introduction, material implication and idempotent are the least common, their are 50 - 150 number of them in P-FOLIO. The occurrences of complex inference rules all range from 1 to 29. The complex inference rules are essential for making certain derivations although their occurrences are less frequent.

\section{Reasoning Evaluation}
We propose three tasks to evaluating LLM reasoning capabilities with \ours at different levels of granularities. 
\subsection{Single-step inference-rule classification}
Given a single step of inference in a human-annotated proof which consists of the premises used $P = \{P_1, P_2\}$ and a derivation $d_1$,  single-step inference-rule classification aims to identify which inference rule has been used to arrive at $d_1$ from $P$. This task aims to evaluate whether a model knows the type of inference required for a single inference step. 
\subsection{Single-step derivation reasoning}
Given a single step of inference in a human annotated proof, the goal of single-step derivation reasoning is to ask the model to generate the truth value of a derivation based on the corresponding premises. The truth value label is always True. 
\subsection{Proof generation}
Given a series of premises $P = \{p_1, p_2,..., p_n\}$ and a conclusion $c_1$, the goal of proof generation is ask the model to generate the proof and the final truth value. This task aims to evaluate the overall proof generation capability.  
\subsection{Proof generation evaluation}
\paragraph{Pass@k.} Considering that a model-generated reasoning chain can diverge significantly from a human-annotated one in logical reasoning, we sample multiple reasoning chains from GPT-3.5 and GPT-4 and employ pass@k metrics \citep{passatk} to assess the quality of these model-generated chains. GPT-4 is utilized to evaluate whether two reasoning chains follow a similar path to reach their conclusions. We provide the prompts used for this in the Appendix. If we sample k proofs from a model, we define pass@k to be the percentage of instances where at least one proof matches the human-written proof.

\subsection{Task comparison}
Single-step inference-rule classification and single-step derivation allow for a more granular assessment of how well the model can follow logical rules and derive conclusions by breaking down complex reasoning chains into individual steps. By analyzing each step independently, identify specific strengths and weaknesses can be identified in the model's reasoning processes, facilitating a better understanding of the model's overall performance and targeted improvements. The entire proof generation evaluates the reasoning capabilities of LLMs by challenging them with complex reasoning problems. This approach requires the model to generate a complete proof from start to finish, demonstrating its ability to handle intricate logical structures and multi-step derivations. By assessing the model's performance on entire proofs, we can gauge its proficiency in maintaining logical coherence, applying relevant rules, and effectively managing the complexity inherent in comprehensive reasoning tasks. This evaluation method provides a holistic view of the model's reasoning abilities.

\section{Experimental Results}
We conduct experiments using P-FOLIO. The train/dev/test split is 70\%/15\%/15\%, which is the same as the split for FOLIO \citep{han2022folio}. For all experiments involved with GPT-4 \citep{openai2023gpt4}, we use gpt-4-0125-preview and the temperature was set to 0. 
\subsection{Single-step evaluation}
We conduct single-step inference rule classification and single-step derivation reasoning on individual proof steps in P-FOLIO. 
For single-step derivation reasoning, we prompt GPT-4 to output the classification results in zero-shot setting and 5-shot prompting. The results of inference rule classification are shown in the first two columns of Table \ref{tab:classification}. Giving GPT-4 a few examples leads to better results for complex inference rules while the improvement is minimal for widely used inference rules. For single-step inference rule classification, we test two settings. We prompt GPT-4 to output the truth value directly in one setting. In the other setting, we prompt GPT-4 to output the explanation first before generating the final truth value. The results of single-step inference rule classification are shown in the last two columns of Table \ref{tab:classification}. Prompting GPT-4 to output the explanation before generating the final truth value leads to better results for both widely-used rules and complex rules.

\begin{table*}[t!]
\centering
\begin{tabular}{>{\raggedright}p{3cm}ccccc}
\toprule
 & \multicolumn{2}{c}{\textbf{Inference Rule Classification}} & \multicolumn{2}{c}{\textbf{Truth Value Classification}}  \\
 & \textbf{No Examples} & \textbf{With Examples}  & \textbf{No Explanation} & \textbf{With Explanation} \\
\midrule
\textbf{Established Rules} & 54.72 & 55.08 & 85.31 & 87.13 \\
\textbf{Complex Rules}      & 40.8 & 43.62 & 70.67 & 80.43   \\
\bottomrule
\end{tabular}
\caption{Single-step inference rule classification and single-step derivation reasoning results. }
\label{tab:classification}
\end{table*}

\subsection{Proof generation}
\paragraph{Zero-shot Prompting LLM} 
We use GPT-4 \citep{openai2023gpt4} to generate proofs and truth value for all the examples. By explicitly prompting GPT-4 to generate the reasoning process before the final answer increase the performance by 5\% over prompting GPT-4 to generate only the final answer. 

\begin{table}[h]
\centering
\begin{tabular}{ccccc}
\toprule
\textbf{\#shot} & \textbf{R-1} & \textbf{R-2} & \textbf{R-L} & \textbf{Acc (\%)} \\ 
\midrule
0-shot & 60.2& 52.7 & 61.5 & 65.2 \\ 
5-shot  & 63.9 & 54.2 & 64.1 & 68.1\\ 
10-shot & 65.8 & 54.9 & 65.9 & 70.7\\ 
20-shot & 70.6 & 59.5 & 69.6 & 75.3\\ 
40-shot & 73.7 & 62.1 & 72.7 & 77.4\\ 
60-shot & 78.2 & 66.3 & 78.1 & 80.6\\ 
\bottomrule
\end{tabular}
\caption{Result of many-shot prompting with GPT-4. }
\label{tab:many-shot-prompting}
\end{table}

\begin{table}[h!]
\centering
\label{tab:dialogue_comparison}
\begin{tabular}{lcc}
\toprule
\textbf{Method} & \textbf{Answer Acc} & \textbf{Proof Acc} \\
w/o proofs. & 60.2 & -  \\ 
w/ proofs. & 65.2 & 58.2 \\ 
\bottomrule
\end{tabular}
\caption{Comparison of final answer accuracy and proof accuracy (\%).}
\end{table}


\begin{table*}[t!]
\centering
\resizebox{\textwidth}{!}{
\begin{tabular}{@{}lcccccccc@{}}
\toprule
\multirow{2}{*} & \multicolumn{2}{c}{FOLIO \citeyearpar{han2022folio}} & \multicolumn{2}{c}{BirdElectricity \citeyearpar{tafjord-etal-2021-proofwriter}} & \multicolumn{2}{c}{ProntoQA \citeyearpar{saparov2023language}} & \multicolumn{2}{c}{bAbi-deductive \citeyearpar{babideductive}} \\
\cmidrule(lr){2-3} \cmidrule(lr){4-5} \cmidrule(lr){6-7} \cmidrule(lr){8-9} 
& w/o Proofs & w/ Proofs & w/o Proofs & w/ Proofs & w/o Proofs & w/ Proofs & w/o Proofs & w/ Proofs \\
Zero-Shot Flan-T5-Large & 5.3\% & 5.3\% & 42.1\% & 40.9\% & 6.5\% & 5.0\% & 38.9\% & 40.4\% \\
Fine-Tune Flan-T5-Large & 65.9\% & 67.5\% & 53.4\% & 60.8\% & 33\% & 41\% & 44.70\% & 50.30\% \\
Zero-Shot Llama3-8B & 52.1\% & 54.1\% & 54.8\% & 57.5\% & 63.1\% & 64.9\% & 72.1\% & 75.7\% \\
Fine-tune Llama3-8B & \textbf{70.2\%} & \textbf{76.3\%} & \textbf{60.8\%} & \textbf{66.7\%} & \textbf{76.1\%} & \textbf{80.6\%} & \textbf{85.1\%} & \textbf{87.9\%} \\
\bottomrule
\end{tabular}
}
\caption{Fine-tuning results and out-of-domain generalization capabilities of LLMs trained on FOLIO and P-FOLIO. "w/o Proofs" indicates the model is trained with FOLIO and "w/Proofs" indicates the model is trained with P-FOLIO.}
\label{tab:out-of-domain}
\end{table*}

We manually inspect all the generated proofs. Among the examples with correct truth values, 10.7\% exhibited incorrect reasoning processes. In these cases, the model either struggled to form the correct reasoning chain for complex problems involving many reasoning steps or used commonsense reasoning as a shortcut. Among the examples with incorrect truth values, 70\% failed to form the correct reasoning chain for complex problems. Additionally, 10\% contained wrong derivations at a certain steps in the reasoning process. About 15\% of the examples presented conclusions with language or logical structures too intricate for the model to fully comprehend. Another 5\% incorrectly applied commonsense reasoning as a shortcut, leading to errors in the final conclusions. We provide more analysis in section \ref{sec:analysis}.

\begin{table}[h!]
\centering
\resizebox{0.45\textwidth}{!}{
\begin{tabular}{lccccc}
\toprule
\textbf{Model} & \textbf{Pass @1} & \textbf{Pass @5} & \textbf{Pass @10} & \textbf{Pass @20} \\
GPT-3.5 & 49.5 & 51.6 & 53.9 & 55.7\\
GPT-4 & 55.3 & 59.2 &  62.1 & 66.9 \\ 
\bottomrule
\end{tabular}
}
\caption{Pass@k results for GPT-3.5 and GPT-4}
\label{tab:passk}
\end{table}

\begin{figure}[!t]
    \centering
    \includegraphics[width=7cm]{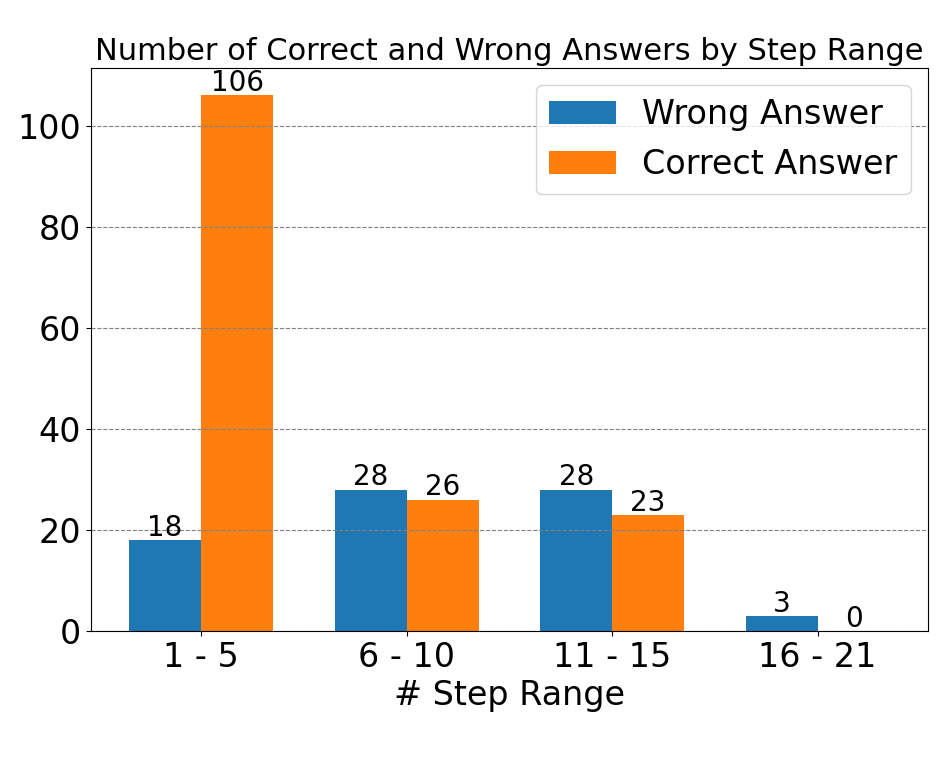}
    \caption{Distribution of the number of correct \& wrong answers vs the number of human-written proof steps}
    \label{fig:error_step_range}
\end{figure}

\begin{table}[h]
\centering
\begin{tabular}{cc}
\toprule
\textbf{Added Rule Type} & \textbf{Accuracy (\%)} \\
\midrule
Complex rules & 70.60 \\
Widely-used rules & 63.20 \\
\bottomrule
\end{tabular}
\caption{Final answer accuracy of GPT-4 with human-written derivations}
\label{tab:rule_accuracy}
\end{table}

\begin{table*}[h]
\centering
\setlength{\tabcolsep}{3pt}
\small
\begin{tabular}{p{15.5cm}}
\toprule
\textbf{Prompt} \\ 
Using deductive reasoning, find out the truth values of the conclusions based on the premises. The truth value can be True, False or Uncertain. First show the reasoning process, and then output the truth value in the format of "Truth value: ".
 Premises: 
 1. If someone has never left the United States, then they have not been to Europe.
 2. If someone has traveled to the Iberian Peninsula, then they have been to Europe.
 3. Everyone who has visited Barcelona has traveled to the Iberian Peninsula.
 4. Everyone who has taken a selfie at the Basilica Familia Sagrada has visited Madrid.
 5. If Kate has either taken a selfie at the Basilica Familia Sagrada or has never left the United States, then Kate has not either visited Barcelona or never left the United States.
  
 Conclusion: 
 Kate has not taken a selfie at the Basilica Familia Sagrada.\\ 
 
\midrule
\textbf{GPT-4 Output} \\ 
Reasoning process:
 Let's analyze the premises and see if we can deduce the truth value of the conclusion.

 Premise 1: If someone has never left the United States, then they have not been to Europe.
 Premise 2: If someone has traveled to the Iberian Peninsula, then they have been to Europe.
 Premise 3: Everyone who has visited Barcelona has traveled to the Iberian Peninsula. Premise 4: Everyone who has taken a selfie at the Basilica Familia Sagrada has visited Madrid. Premise 5: If Kate has either taken a selfie at the Basilica Familia Sagrada or has never left the United States, then Kate has not either visited Barcelona or never left the United States.
... \textit{Verbatim repetition of the original premises...}

 Conclusion: Kate has not taken a selfie at the Basilica Familia Sagrada.

 From Premise 5, we know that if Kate has taken a selfie at the Basilica Familia Sagrada, then she has visited Barcelona or left the United States. However, we don't have any information about whether Kate has actually taken a selfie at the Basilica Familia Sagrada or not. Therefore, we cannot deduce the truth value of the conclusion based on the given premises.
 
 Truth value: Uncertain \\
\bottomrule
\end{tabular}
\caption{Case Study}
\label{tab:case-study}
\end{table*}

\paragraph{Many-shot prompting LLMs}
We conduct many-shot prompting experiments with GPT-4 using 5, 10, 20, 40, and 60 examples in in-context learning. Sixty is the maximum number of example and proof pairs we can fit into the context window of GPT-4. We present the results using ngram-overlap-based metrics, including ROUGE-1, ROUGE-2, and ROUGE-L \citep{lin-2004-rouge}, comparing the model-generated proofs with human-written proofs. Additionally, we report the final answer prediction accuracy. Table \ref{tab:many-shot-prompting} demonstrates that adding 5 examples to the prompt improves GPT-4's performance over zero-shot prompting. Increasing the number of examples for in-context learning from 5 to 20 results in a further performance improvement of approximately 7\%. Using 60 examples in the prompt significantly enhances performance compared to zero-shot prompting.

\paragraph{Pass@k Evaluation}
Table~\ref{tab:passk} presents the pass@k results for GPT-3.5 and GPT-4 on the P-FOLIO dataset, illustrating the comparative performance of these models in generating reasoning chains. Pass@k metrics indicate the percentage of instances where at least one out of k sampled reasoning chains matches the human-annotated proof. The results show a clear performance improvement from GPT-3.5 to GPT-4 across all values of k. Specifically, GPT-4 achieves a pass@1 of 55.3\%, outperforming GPT-3.5's 49.5\%. This trend continues with GPT-4 attaining pass@5, pass@10, and pass@20 scores of 59.2\%, 62.1\%, and 66.9\% respectively, compared to GPT-3.5's 51.6\%, 53.9\%, and 55.7\%. These findings suggest that GPT-4 is more effective in generating reasoning chains that align with human-annotated proofs, highlighting its enhanced capability in logical reasoning tasks.

\paragraph{Fine-tuning and Out-of-Domain Generalization}
We test the performance of fine-tuning with Flan-T5-Large \citep{flan} and Llama3-8B \citep{llama3modelcard} and test the out-of-domain generalization capabilities of LLMs fine-tuned on P-FOLIO. We compares two models, Flan-T5-Large and Llama3-7B \citep{llama3modelcard}, in both zero-shot and fine-tuned configurations. We provide the hyper-parameter settings in the Appendix. 
Table~\ref{tab:out-of-domain} highlights that fine-tuning on FOLIO (w/o Proofs) and P-FOLIO (w/Proofs) significantly enhances the performance of both models across all tasks. Llama3-7B consistently outperforms Flan-T5-Large. 

Table~\ref{tab:out-of-domain} also shows that fine-tuning without proofs using the FOLIO dataset and fine-tuning with proofs boost the performance on the three out-of-domain logical reasoning datasets tested, BirdElectricity \citep{tafjord-etal-2021-proofwriter}, ProntoQA \citep{saparov2023language} and bAbi-deductive \citep{babideductive}. The inclusion of human-written proofs consistently improves the accuracy for fine-tuned models, underscoring the importance of proofs in bolstering the generalization capabilities of LLMs in out-of-domain datasets. These results suggest that LLMs fine-tuned with human-written rationales are able to generalize learned reasoning strategies to new, unseen tasks, demonstrating robust out-of-domain generalization. 

\section{Analysis}
\label{sec:analysis}
\paragraph{Most powerful LLMs fail on examples with 10 steps or more in human-written reasoning chains.}

Figure~\ref{fig:error_step_range} illustrates the performance of GPT-4 on logical reasoning tasks of varying complexity. While GPT-4 accurately answers most questions requiring 1-5 human-written reasoning steps, its accuracy drops significantly for more complex tasks. For questions involving 6-10 reasoning steps, GPT-4 answers correctly only about half the time. Its performance further declines with examples requiring 11-15 steps, where it fails more than half the time. For tasks with 16 or more reasoning steps, GPT-4 is unable to provide correct answers at all.

\paragraph{Making new derivations with complex rules could be one of the bottlenecks for LLM reasoning}
To gain a more detailed understanding of LLM reasoning capabilities, we conducted an experiment by augmenting the original logical reasoning examples with additional derivations. First, we created a set of examples by adding derivations derived exclusively from widely-used rules, ensuring that no complex or less common rules were included. Next, we created another set of examples by incorporating derivations that were based solely on well-established rules, explicitly excluding any derivations that originated from widely-used rules. This approach allowed us to isolate the impact of different types of inference rules on the reasoning performance of LLMs. 

Table~\ref{tab:rule_accuracy} reports the zero-shot prompting performance of GPT-4 on the two new set of augmented examples we created. The results reveal that GPT-4 achieves much higher accuracy on the set augmented with complex inference rule derivations, at 70.60\%, compared to an accuracy of 63.20\% on the set with widely-used rule derivations. This indicates that while GPT-4 performs better when leveraging complex inference rules, the creation of new derivations using these rules could pose a significant challenge, potentially acting as a bottleneck for large language model (LLM) reasoning. This suggests that further advancements in handling complex rule derivations may be necessary to improve the overall reasoning capabilities of LLMs.

\paragraph{Faulty reasoning}
We show a case study where GPT-4 makes a mistake in writing the proof and producing the final truth value. The logical reasoning example and the GPT-4 output are shown in Table~\ref{tab:case-study}. This mistake happens due to the model's inability to form the correct reasoning chain for a complex problem involving lots of steps, which is also the main cause of most model failures. Notably, in this scenario, the model indeed follows formal rules in each step perfectly and can make correct derivations based on the premises provided. However, it is unable to reach a conclusion because it is unable to correctly derive certain crucial intermediate steps in the reasoning process. We observe a similar pattern across different examples that require 10 reasoning steps or more. As we can see from the GPT-4 output, the model fails to find the correct reasoning chain, thereby not finding the final conclusion and producing "Uncertain" for the truth value although the correct answer should be "True". In particular, the model fails to arrive at the derivation that "No one both has visited Ibiza and never left the United States.", which can be derived from "If someone has visited Ibiza, then they have left the United States." using one of the complex rules:  "from $a \rightarrow \neg b$, we know $\lnot (a \land b)$".

\section{Conclusion}
We presented P-FOLIO, a new dataset consisting of human-written proofs for FOLIO, a set of realistic logical reasoning stories written by humans. $26\%$ of a total of 1430 proofs in \ours consist of more than five reasoning steps and $4\%$ of proofs consist of more than 10 reasoning steps. We further use \ours to evaluate and improve large-language-model reasoning capabilities. We propose tasks with different levels of granularities and use various metrics to evaluate the performance of LLM reasoning capabilities. We show that human-written reasoning chains significantly boost the logical reasoning capabilities of LLMs via many-shot prompting and fine-tuning. Fine-tuning on P-FOLIO improves the model performance by more than 10\% on three other out-of-domain logical reasoning datasets. We conduct detailed analysis to show where most poweful LLMs fall short in reasoning.


\section{Limitations}
The abundant human-written reasoning chains in \ours can be used in many other ways that we have not explored. For instance, bootstrapping with abundant human-written reasoning chains could further improve the performance of bootstrapping with limited human-written rationales \citep{zelikman2022star}. 

\bibliography{anthology,custom}
\bibliographystyle{acl_natbib}

\appendix

\section{Appendix}
\label{sec:appendix}
\subsection{Annotation Guidelines}
We provide the detailed annotation guidelines in Table~\ref{tab:annotation_guidelines}.

\begin{table*}[h]
\centering
\setlength{\tabcolsep}{3pt}
\small
\begin{tabular}{p{15.5cm}}
\toprule
Write a proof for each conclusion in a story. Write the relevant derivation at each inference step and the rule of inference.

Please think about the forms of the input premises and conclusions carefully. If the forms do not correspond to a rule in the instructions, then you should not use it without verifying it.
You can invent new rules yourselves. 
But please verify the correctness of the rules with the truth table calculator \href{https://www.emathhelp.net/calculators/discrete-mathematics/truth-table-calculator/}{here}. If the rule is not successfully verified, please do not use it. If it is successfully verified, you can add it to the updated list and use it. 

\\
\textbf{Do’s:} \\ 
\textbf{***} Write the inference rule used for each inference step.
For inference rules, please refer to this \href{https://en.wikipedia.org/wiki/List_of_rules_of_inference#Rules_for_negations}{page}.   If you think the inference rules needed are not included in this list, please add them to the updated list below. \\ 
... \textit{Established inference rule definitions with examples}...
\\ 
\textbf{Intuitively and Logically Correct Rules}: There are some rules that might be needed in the proof procedure but that are hard to be derived. Theses rules do not have a commonly agreed name, we are listing these rules as below: \\

\textbf{***} $\lnot(a \xor b)$ is equivalent to $(\lnot a \land \lnot b)$ or $\lnot (a \land b)$. \\ 

\textbf{***} ANDEquivalent1: $a \rightarrow \lnot b$ is equivalent to $\lnot(a \land b)$ \\ 

\textbf{***} XOR1: $\lnot(a \xor \lnot b)$ is equivalent to $(a \xor b)$ \\ 

\textbf{***} XOR2: $(a \xor b)$ is equivalent to $(\lnot a \xor \lnot b)$ \\

\textbf{***} XOR3: $(\lnot b \xor b)$ is always true. \\ 

\textbf{***} XOR4: From $(a \xor b)$, we know that $a \rightarrow \lnot b$.  \\

\textbf{***} XOR5: From $\lnot(a \xor b)$, we know that $a \rightarrow b$. \\

\textbf{***} XOR6: From $a \rightarrow b$ and $b \rightarrow a$, we know that $\lnot(a \xor b)$.  \\

\textbf{***} IMPAND1: From $a \rightarrow \lnot b$, we know that $\lnot (a \land b)$.

\textbf{***} XORUni1: From $\forall x (C(x) \rightarrow \lnot A (x))$, $\forall x (A(x) \xor B(x))$, we know that $\forall x (C(x) \rightarrow B(x))$. \\ 

\textbf{***} XORUni2: From $\forall x (A(x) \rightarrow B(x))$, $\forall x (A(x) \xor B(x))$, we know that $\forall x (B(x))$.  \\

\textbf{***} XORUni3: From $\forall x (A(x) \xor B(x))$ and $\forall x (A(x) \rightarrow C(x))$, we know that $\forall x (C(x) \lor B(x))$. \\

\textbf{***} OREquivalent1: $ (a \lor b) \land (a \lor \lnot b)$ is equivalent to $a$ \\ 

\textbf{***} ANDUni1: From $\forall x (A(x) \rightarrow C(x))$ and $A(paul) \land B(paul)$”, we know that "$C(paul) \land B(paul)$". \\

\textbf{***} ORUni1: From $\forall x (A(x) \lor B(x))$ and $\forall x (A(x) \rightarrow C(x))$, we know that $\forall x (C(x) \oplus B(x))$.

\textbf{***} ORUni2: From $\forall x (A(x) \rightarrow C(x))$, $A(\text{paul}) \lor B(\text{paul})$, we know that $C(\text{paul}) \lor B(\text{paul})$.

\textbf{***} ORUni3: From $\forall x (\neg C(x) \rightarrow A(x))$, $A(\text{paul}) \lor B(\text{paul})$, we know that $\neg C(\text{paul}) \lor B(\text{paul})$.

\textbf{***} ORUni4: From $\forall x (A(x) \rightarrow C(x))$ and $\forall x (B(x) \rightarrow C(x))$, we know that $\forall x ((A(x) \lor B(x)) \rightarrow C(x))$.

\textbf{***} Note: From $\neg a \rightarrow b$, we cannot know $\neg (a \land b)$.

\textbf{***} Note: From $\forall x (A(x) \oplus B(x))$ and $\forall x (A(x) \rightarrow C(x))$, we cannot infer $\forall x (C(x) \oplus B(x))$.

\textbf{***} Also note: From $\forall x (A(x) \rightarrow D(x))$ and $A(\text{paul}) \lor B(\text{paul}) \rightarrow C(\text{paul})$, we cannot infer $D(\text{paul}) \lor B(\text{paul}) \rightarrow C(\text{paul})$.

\textbf{***} ORUni5: From $\forall x (C(x) \rightarrow \neg A(x))$, $\forall x (A(x) \lor B(x))$, we know that $\forall x (C(x) \rightarrow B(x))$.

\textbf{***} Iff introduction: If A, then B. If B, then A. Then, either both A and B or neither A nor B. \\

\textbf{***} For complicated conclusions, sometimes proving part of them is sufficient. A or B. A is true -> the conclusion is true.
A -> B. A is false -> the conclusion is false. 
For conclusions with truth values of False, prove the opposite is True.  \\

\textbf{***} For the sake of annotation purposes, either..or in any context is exclusive, or is inclusive. For example, if I can either sleep at home today or go to class, this means that I can choose to sleep at home, go to class, but not both. Whereas if I can I can drink soda or drink coffee, it means I can drink soda, coffee, or both. \\
\textbf{***} When we say that someone is not an energetic person or a sloth, this means that someone is neither an energetic person nor a sloth. \\
\textbf{***} $\neg (\text{Indian}(\text{jesse}) \oplus \text{Human}(\text{jesse}))$
 : Jessie is either Indian and human, or that he is neither Indian nor human. In several instances, this sentence is rendered as "Jessee is not either an Indian or a human."  \\
\\ 
\textbf{Proof consistency check:} \\ 
\textbf{***} Please involve at most two premises in one inference step. \\
\textbf{***} Be sure to write the conjunction introduction steps. \\
\textbf{***} Existential generalization check: Broadway Sheetmetals was a business owned by Edwin Smith, who was a rower. -> There was a business owned by a rower. There are two steps of existential generalization involved. Be sure to write both steps. \\
\textbf{***} Do not use derivations used for a previous conclusion in a new conclusion. If conclusion A uses certain inference steps of conclusion B you have written the proof for, be sure to copy those inference steps in the proof of conclusion A. \\
\textbf{***} Check for missing modus tollens steps.  \\ 
\textbf{***} Note the difference between modus tollens and transposition.\\
\textbf{***} Note the difference between hypothetical syllogism, modus ponens and universal instantiation. \\
\textbf{***} Hypothetical syllogism should be easy to distinguish from the other two since hypothetical syllogism would involve two implications (->). Both the premises need to have implications in order to derive the conclusion. \\
\textbf{***} Universal instantiation and modus pones both involve only one implication \\ 
\textbf{***} Universal instantiation involves the universal quantifier, you can check the FOL part if you are not sure if a statement involves the universal quantifier. \\
\textbf{***} Modus ponens does not involve the universal quantifier. \\
\textbf{***} Note: A(a) xor B(a) imply A(a) or B(a); A(a) or B(a) cannot imply A(a) xor B(a).  \\
\textbf{***} Hypothetical syllogism instead of universal instantiation: If you have room for dessert, you have room for broccoli. Everyone at Luis's dinner party has room for dessert, including Luis. -> Everyone at Luis's dinner party has room for broccoli.  \\

\bottomrule
\end{tabular}
\caption{Annotation guidelines}
\label{tab:annotation_guidelines}
\end{table*}

\subsection{Instruction used in the GPT-4 prompt}
Table~\ref{tab:proof_gen_prompt} shows the instruction used in the GPT-4 prompt.
\begin{table*}[t!]
    \centering
    \begin{tabularx}{\textwidth}{X}
    \toprule
        Using deductive reasoning, find out the truth values of the conclusions based on the premises. The truth value can be True, False or Uncertain. First show the reasoning process, and then output the truth value in the format of "Truth value: ".\\
    \bottomrule
    \end{tabularx}
    \caption{Instruction used for prompting GPT-4 for zero-shot proof generation.}
    \label{tab:proof_gen_prompt}
\end{table*}

\subsection{Prompt used for evaluating  reasoning path}
We provide an example in Table~\ref{tab:case-study-gpt4-compare-two} including our prompt used and GPT-4 input and output for how GPT-4 is used to evaluate whether two reasoning chains follow a similar path to reach their conclusions.

\begin{table*}[h]
\centering
\setlength{\tabcolsep}{3pt}
\small
\begin{tabular}{p{15.5cm}}
\toprule
\textbf{Prompt} \\ 
\textit{Given a deductive reasoning question, demonstrate whether the two reasoning chains are semantically similar and follow the same reasoning path to derive the final answer. After your explanations, output your decision in the format of "Decision: ". Your decision should be either Yes or No.} \\ 
\textbf{Premises}: 
 1. Northwich Victoria FC is a football club that plays at Wincham Park.
 2. Simon Hackney is a member of Northwich Victoria FC.
 3. Northwich Victoria FC is part of the Northern Premier League.
 4. If someone is a member of an organization that is part of a group, then the person is a member of the group.
 5. If someone is a member of an organization and the organization plays at a certain place, then the person plays at the place.
  
\textbf{Conclusion}: 
Simon Hackney is a member of the Northern Premier League.

\textbf{Reasoning chain A:} 
Reasoning process:
 1. Simon Hackney is a member of Northwich Victoria FC (Premise 2).
 2. Northwich Victoria FC is part of the Northern Premier League (Premise 3).
 3. If someone is a member of an organization that is part of a group, then the person is a member of the group (Premise 4).
 4. Therefore, Simon Hackney is a member of the Northern Premier League (from 1, 2, and 3).
 
\textbf{Reasoning chain B}: 
Reasoning process: 
From premise 2 and premise 3, we know that Simon Hackney is a member of Northwich Victoria FC, which is a part of the Northern Premier League. 
From this and premise 4, we know that Simon Hackney is a member in the Northern Premier League. \\ 
\midrule
\textbf{GPT-4 Output} \\ 
\textbf{Analyzing Reasoning Chain A}:
Step 1: It asserts that Simon Hackney is a member of Northwich Victoria FC, which directly uses Premise 2.
Step 2: It states that Northwich Victoria FC is part of the Northern Premier League, utilizing Premise 3.
Step 3: It invokes the rule from Premise 4, which is a conditional statement about membership extension from an organization to a group.
Step 4: This is the conclusion step that synthesizes the information from steps 1, 2, and 3 to conclude that Simon Hackney is a member of the Northern Premier League. This follows a logical and direct application of the premises.
\textbf{Analyzing Reasoning Chain B}:
Initial Statement: Acknowledges that Simon Hackney is a member of Northwich Victoria FC (from Premise 2) and that the club is part of the Northern Premier League (from Premise 3).
Subsequent Logic: Directly applies Premise 4, which relates individual membership in an organization to membership in a broader group if the organization is part of that group.
Conclusion: Concludes that Simon Hackney is a member of the Northern Premier League, which aligns with the conclusion from the initial statement and the application of Premise 4.
Comparison and Decision:
\textbf{Both reasoning chains follow the same logical structure:}
They acknowledge Simon Hackney's membership in Northwich Victoria FC.
They recognize the club’s affiliation with the Northern Premier League.
They apply the same general rule from Premise 4 to extend Simon's membership from the club to the league.
The primary difference lies in the explicitness and breakdown of steps:
Chain A is more structured and delineated, breaking down each logical step explicitly.
Chain B is more concise, combining some steps but still following the same logical pathway.
\textbf{Decision: Yes — Both reasoning chains are semantically similar and follow the same reasoning path to derive the final answer.}\\ 
\bottomrule
\end{tabular}
\caption{Example including prompt, GPT-4 model input and output for evaluating whether two model outputs following the same reasoning path.}
\label{tab:case-study-gpt4-compare-two}
\end{table*}

\subsection{Hyper-parameter settings}
For fine-tuning Flan-T5-Large, we use a learning rate of 1e-4, batch size of 8, gradient accumulation of 2, warmup step of 100 and the model was trained for 3 epochs. For fine-tuning LLama-8b, we use a learning rate of 2e-5, batch size of 1 and gradient accumulation of 8. The warmup steps were set to 25 and the model was trained for 3 epochs. 


\end{document}